\documentclass{article}

\usepackage[english]{babel}

\usepackage[letterpaper,top=2cm,bottom=2cm,left=3cm,right=3cm,marginparwidth=1.75cm]{geometry}

\usepackage{tabularx}  
\usepackage{subcaption}
\usepackage{amsmath}  
\usepackage{amssymb}  
\usepackage{graphicx}
\usepackage{authblk}
\usepackage{float}
\usepackage{booktabs}
\usepackage{csquotes}
\usepackage{tikz}
\usepackage[style=numeric,sorting=none]{biblatex}
\usepackage[colorlinks=true, allcolors=blue]{hyperref}

\title{SwinECAT: A Transformer-based fundus disease classification model with Shifted Window Attention and Efficient Channel Attention}

\author[1]{Peiran Gu}
\author[1]{Teng Yao}
\author[2]{Mengshen He}
\author[1]{Fuhao Duan}
\author[1]{Feiyan Liu}
\author[1]{RenYuan Peng}
\author[1]{Bao Ge}
\affil[1]{School of Physics and Information Technology, Shaanxi Normal University, Xi’an, China}
\affil[2]{School of Computer Science and Engineering, Central South University, Changsha, China}
\begin{document}
\maketitle

\begin{abstract}
In recent years, artificial intelligence has been increasingly applied in the field of medical imaging. Among these applications, fundus image analysis presents special challenges, including small lesion areas in certain fundus diseases and subtle inter-disease differences, which can lead to reduced prediction accuracy and overfitting in the models. To address these challenges, this paper proposes the Transformer-based model SwinECAT, which combines the Shifted Window (Swin) Attention with the Efficient Channel Attention (ECA) Attention. SwinECAT leverages the Swin Attention mechanism in the Swin Transformer backbone to effectively capture local spatial structures and long-range dependencies within fundus images. The lightweight ECA mechanism is incorporated to guide the SwinECAT's attention toward critical feature channels, enabling more discriminative feature representation. In contrast to previous studies that typically classify fundus images into 4 to 6 categories, this work expands fundus disease classification to 9 distinct types, thereby enhancing the granularity of diagnosis. We evaluate our method on the Eye Disease Image Dataset (EDID) containing 16,140 fundus images for 9-category classification. Experimental results demonstrate that SwinECAT achieves 88.29\% accuracy, with weighted F1-score of 0.88 and macro F1-score of 0.90. The classification results of our proposed model SwinECAT significantly outperform the baseline Swin Transformer and multiple compared baseline models. To our knowledge, this represents the highest reported performance for 9-category classification on this public dataset.
\end{abstract}


\section{Introduction}

Currently, various fundus diseases have become a significant concern impacting both individuals' quality of life and broader public health. These conditions not only severely impair patients' vision and daily functioning but also impose substantial burdens on healthcare systems (~\cite{williams2022addressing,elam2022disparities}) . Fundus imaging has become an essential tool in ocular disease diagnosis, offering high-resolution visualization of retinal structures. It plays a critical role in the diagnosis of a wide range of ocular pathologies (~\cite{ting2019artificial,gulshan2016development}) . The imaging principle of fundus photography is based on illuminating the fundus with incident light and capturing the reflected light on a photosensitive imaging plane, thereby converting the three-dimensional anatomical structure of the fundus into a two-dimensional image (\cite{gupta2024comprehensive}) . With the development of medical imaging technology and artificial intelligence, deep learning methods are gradually being used in the diagnosis of fundus diseases (~\cite{azad2024advances,zhang2019automated,zuo2025machine}) .

Visual models based on the Transformer architecture have been extensively applied in medical image analysis (~\cite{azad2024advances}). Compared to convolutional neural networks (CNNs), Transformer models offer several advantages: their global modeling capability allows better capture of long-range dependencies, making them well-suited for tasks involving irregular lesion distributions; furthermore, their multi-scale feature processing facilitates the identification of lesions of varying sizes (~\cite{shamshad2023transformers,takahashi2024comparison,cheng2024review}). Previous studies have frequently adopted model combination approaches to improve classification performance. For example, Liu et al. proposed a feature fusion model that integrates MaxViT and ResNet (~\cite{liu2024automated}) , while Lian et al. proposed a model based on the sequential connection of CNN and Transformer (\cite{lian2024lesion}) . However, these approaches substantially increase the number of model parameters, which may lead to overfitting, reduce model generalizability, and ultimately impair overall performance. Most existing research is limited to assessing the severity of a single fundus condition (~\cite{gao2015automatic,gao2018diagnosis,gao2018diagnosis,kim2024deep}) or handling three to six diagnostic classes (~\cite{hasan2025dia,albelaihi2024deepdiabetic,ranjith2025novel}) . Furthermore, a distinctive feature of fundus images is that many lesions manifest as small targets, such as hemorrhagic spots and other abnormalities, which are difficult to detect and impose stringent requirements on model design and selection (~\cite{gour2023challenges,abdullah2024review,bernabe2021classification}) .

Therefore, we propose SwinECAT (Shifted Window-based Efficient Channel Attention Transformer) , an improved model that combines the Shifted Window (Swin) Transformer architecture (~\cite{liu2021swin}) with Efficient Channel Attention (ECA) (~\cite{wang2020eca}) . Specifically, SwinECAT progressively constructs multi-scale feature maps through a four-stage hierarchical architecture, with each stage integrating the Swin Attention and the ECA. The backbone of Swin Transformer has the capacity to effectively capture both local details and global structures. Its window-based self-attention mechanism facilitates efficient extraction of retinal lesion features within localized regions, while the shifted window strategy enables information exchange across windows, supporting the modeling of spatial dependencies in fundus images. The ECA module is based on the lightweight attention, emphasizes useful feature channels, and has high computational efficiency (~\cite{wang2020eca}) . We incorporate it into the backbone of Swin Transformer to enhance the feature channel expression ability of the backbone of Swin Transformer model (~\cite{liu2021swin,wang2020eca}) , making the network pay more attention to the information of the main channel, which helps to increase the feature selection during model training, is more sensitive to small target recognition in medical images, and improves the model classification ability. In addition, because ECA does not use the fully connected layer and only uses 1D convolution to achieve channel interaction, SwinECAT has a small parameter increment in the backbone of Swin Transformer.

To evaluate the effectiveness of our proposed SwinECAT model, this study analyzes the characteristics of several widely used baseline models in general vision tasks (~\cite{chen2022advance}) and selects six representative models for comparative experiments. These include five Transformer-based models: ViT (\cite{dosovitskiy2020image}), ConViT (\cite{d2021convit}), BEiT (\cite{bao2021beit}), MaxViT (\cite{tu2022maxvit}), and Swin Transformer (\cite{liu2021swin}), as well as one CNN-based model: ResNet50 (\cite{he2016deep}). In addition, the proposed SwinECAT model is compared with 2 recently published baseline models designed for fundus disease classification: the feature fusion model of MaxViT and ResNet (\cite{liu2024automated}) and the sequential connection model of CNN and Transformer (\cite{lian2024lesion}). Experimental results demonstrate that SwinECAT achieves the highest accuracy among these models while maintaining a lower parameter count.

Compared with previous related studies, this paper has the following contributions:

\hspace*{2em}1) We propose the SwinECAT model. In terms of model improvement, previous studies on fundus disease image classification usually used model combination methods to improve classification performance. For example, recently proposed methods include the feature fusion model of MaxViT and ResNet (~\cite{liu2024automated}) , and the sequential connection model of CNN and Transformer (~\cite{lian2024lesion}) . Different from previous studies, Our SwinECAT model integrates both spatial and channel attention mechanisms, and demonstrates superior performance over previous improved methods in terms of both parameter efficiency and classification accuracy.  In order to improve accuracy while reducing the high enhancement of model parameters, this paper adopts a lightweight Efficient Channel Attention mechanism and improves the backbone of Swin transformer with it, so that the classification performance of the model is improved with only a small increase in model parameters. The SwinECAT model we designed achieved an accuracy of 88.29\% on the Eye Disease Image Dataset (EDID) (~\cite{sharmin2024dataset}) , which is higher than several widely used baseline models in general vision tasks, and also outperforms 2 recently published models designed for fundus disease classification: the feature fusion model of MaxViT and ResNet (~\cite{liu2024automated}) and the sequential connection model of CNN and Transformer (~\cite{lian2024lesion}) .

\hspace*{2em}2) Most previous studies focus on analyzing the severity of a single fundus disease  (~\cite{gao2015automatic,gao2018diagnosis,gao2018diagnosis,kim2024deep}) or distinguishing between fewer than nine diagnostic categories (~\cite{hasan2025dia,albelaihi2024deepdiabetic,ranjith2025novel}) . Compared with previous studies, our study expands the comprehensiveness of deep learning in fundus disease classification. The EDID dataset selected in this paper contains 8 types of diseases and healthy fundus images (~\cite{sharmin2024dataset}) . Fundus diseases are characterized by small targets. Increasing the number of diseases poses a great challenge to the classification ability of the model (~\cite{gour2023challenges,abdullah2024review}) . However, The model SwinECAT still achieves a high classification accuracy on this dataset.

The rest of this paper is as follows: Section 2 introduces related research and several baseline models for general vision tasks; Section 3 introduces the SwinECAT model and its key modules; Section 4 presents the experimental results and discusses the experiments, including comparison between SwinECAT and some Transformer-based visual baseline model, comparison between SwinECAT and the baseline models proposed by previous studies, ablation experiments, and the discussion of the experimental process; Section 5 concludes our research.

\section{Related work}

This section first reviews existing deep learning-based approaches for fundus disease image classification, followed by an introduction to several widely adopted baseline models in general vision tasks, which establish the foundation for the proposed method.

\subsection{Deep learning models for fundus disease image classification}

\subsubsection{CNN-based deep learning models}

Before the Convolutional neural networks (CNNs) model are used to classify funds disease images, early methods for analyzing fundus diseases are mainly based on manual methods, such as manual feature extraction and grading or classification (~\cite{age2001amd,age2001cataract,roychowdhury2013dream}) , which are difficult to use on a large scale.

CNNs have significant advantages over traditional machine learning methods (such as SVM, KNN, etc.) . For example, CNNs can automatically extract multi-level features from original images, avoiding the limitations of manual feature extraction and having stronger expression capabilities and classification accuracy (~\cite{krizhevsky2012imagenet,litjens2017survey,abdulwahhab2024review}) . Neural network-based models are used in the medical field to diagnose various diseases due to their powerful feature extraction and high-precision classification capabilities, including remarkable achievements in the field of fundus disease classification. Gao et al. used convolutional neural networks and recurrent neural networks to grade nuclear cataracts (~\cite{gao2015automatic}) . Gao et al. used ResNet, VGG and other neural networks to achieve automatic diagnosis of diabetic retinopathy (~\cite{gao2018diagnosis}) . Zhang et al. proposed the DeepDR model, which can be used for automatic identification and grading of diabetic retinopathy. DeepDR detects the presence and severity of DR directly from fundus images through transfer learning and ensemble learning (~\cite{zhang2019automated}) .

Although CNN-based fundus disease classification methods have achieved a lot of results, there are still some problems. For example, CNN-based models are good at local convolution features and it is difficult to consider long-distance pixel dependencies like Transformer-based visual models. The overall receptive field of CNN-based models in images is not as good as the global modeling ability of Transformer-based visual models. (~\cite{bali2024analysis,salehi2023study,alzubaidi2021review}) .

We note that most studies based on CNN deep learning models focus on the severity analysis of a single fundus disease (~\cite{zhang2023application,foo2021artificial,gao2015automatic,gao2018diagnosis}) . This is because fundus disease images have the characteristics of small lesion features, and small differences between categories (~\cite{gour2023challenges,abdullah2024review}) , which places high demands on the selection and improvement points based on deep learning models.

\subsubsection{Attention-based deep learning models}

The Transformer model was proposed by Google in 2017. It is a new network architecture based on the attention mechanism (~\cite{vaswani2017attention}) . It performs well in machine translation and has a profound impact on the field of computer vision. The Transformer-based visual models are widely used in the fields of computer vision and medical imaging due to its powerful global modeling capabilities, more flexible feature expression capabilities and greater suitability for large-scale training (~\cite{azad2024advances}) . Liu et al. classified nine fundus diseases in infants based on residual neural network and MaxViT feature fusion method (~\cite{liu2024automated}) . Lian et al. proposed a CNN-Transformer structure model, which first extracts the features of fundus images through CNN, and then uses the output of CNN as the input of the subsequent visual Transformer to assist in the diagnosis of diabetic retinopathy (~\cite{lian2024lesion}) . D. Ranjith et al. combined attention mechanism, late fusion and early fusion with graph neural network to classify cataract, glaucoma and normal (~\cite{ranjith2025novel}) .

Compared with CNN, which relies on local convolution operations, the Transformer-based visual models capture the long-distance dependencies between image regions through the global self-attention mechanism (~\cite{krizhevsky2012imagenet,vaswani2017attention,dosovitskiy2020image}) . It is suitable for identifying the overall spatial layout of fundus images and coping with the challenges of multi-scale lesions. Research on fundus disease classification based on the Transformer-based visual models has gradually outperformed the models based on CNN alone without relying on the attention mechanism in terms of data scale, such as data quantity, data type, and classification accuracy (~\cite{wang2024multi,liu2024automated}) .

\subsection{Several widely used baseline models for general vision tasks}

ResNet is a deep CNN architecture proposed by He et al. It effectively solves the common gradient vanishing and degradation problems in deep network training by introducing residual connections (~\cite{he2016deep}) . It has achieved excellent performance in multiple general vision tasks. In recent years, ViT and its improved models have achieved remarkable results in the field of computer vision. ViT first applied Transformer to image classification tasks (~\cite{dosovitskiy2020image}) . Subsequently, models such as Swin Transformer (~\cite{liu2021swin}) , ConViT (~\cite{d2021convit}) , BEiT (~\cite{bao2021beit}) , MaxViT (~\cite{tu2022maxvit}) were proposed one after another. They were optimized from the perspectives of model structure, computational efficiency, pre-training method, attention mechanism, etc., and performed well in different task fields. The characteristics and advantages of the models are shown in the following table \ref{tab:vit_models}:

\begin{table}[H]
\centering
\begin{tabular}{l p{10cm}} 
\toprule
\multicolumn{1}{c}{Models} & \multicolumn{1}{c}{Characteristics and advantages} \\
\toprule
\multicolumn{1}{c}{ResNet} (~\cite{he2016deep})               & It enables deep network training through residual connections, enhancing performance on vision tasks.                          \\
\multicolumn{1}{c}{ViT} (~\cite{dosovitskiy2020image})               & Transformer is successfully applied to vision tasks for the first time, achieving end-to-end image classification.                          \\
\multicolumn{1}{c}{Swin Transformer} (~\cite{liu2021swin})  & By combining local window attention with the shifted window strategy, it achieves both effective local modeling and enhanced global perception.                                 \\
\multicolumn{1}{c}{ConViT} (~\cite{d2021convit})            & Combining convolution and self-attention mechanisms improves the model's ability to capture local information.                               \\
\multicolumn{1}{c}{BEiT} (~\cite{bao2021beit})              & Through self-supervised pre-training, the introduction of visual vocabulary significantly improves the performance of downstream tasks.      \\
\multicolumn{1}{c}{MaxViT} (~\cite{tu2022maxvit})            & A hybrid model that integrates convolution and multiple attention mechanisms of Transformer to achieve efficient local-global feature extraction.                                                 \\
\bottomrule
\end{tabular}
\caption{Characteristics and advantages of several widely used baseline models for general vision tasks}
\label{tab:vit_models}
\end{table}

\section{Materials and methods}

This section primarily describes the dataset source and preprocessing steps, our proposed SwinECAT model, and the evaluation metrics used. The core focus of this section is the SwinECAT model, which is detailed in terms of its overall architecture and key components.

\subsection{Dataset}

\subsubsection{Dataset description}

The experimental data used in this study comes from the Eye Disease Image Dataset (EDID) (~\cite{sharmin2024dataset}) released in 2024. The dataset was jointly constructed by Anawara Hamida Eye Hospital and B.N.S.B. Zahurul Haque Eye Hospital in Faridpur, Bangladesh. Fundus images were collected using two Topcon digital fundus cameras (models: TRC-50DX and TL-211).

The EDID augmented dataset contains a total of 16,242 high-quality images with the image size of 2004×1690 pixels, covering clinical samples of healthy people and nine representative fundus lesions. This dataset contains fundus images of 8 types of fundus diseases, fundus images of healthy eyes, and external ocular images of pterygium. Because pterygium data does not belong to fundus images, it is excluded in the research of this paper. That is, this article uses a total of 16,140 fundus images, including 8 types of disease and health data. A brief description of 8 types of fundus diseases and healthy is shown in Table \ref{tab:various fundus diseases}:

\begin{table}[H]
\centering
\begin{tabular}{@{}
  >{\centering\arraybackslash}p{4cm} 
  >{\centering\arraybackslash}p{4cm} 
  p{8cm}
@{}}
\toprule
\textbf{Healthy and types of fundus diseases} & \textbf{Number of images} & \textbf{Brief description of healthy and fundus disease} \\
\midrule
Healthy                          &    2676                       & Normal fundus appearance without significant pathological changes \\
Retinitis Pigmentosa             &    834                       & A hereditary, progressive degenerative disease of retinal photoreceptor cells (\cite{hartong2006retinitis}) \\
Retinal Detachment               &    750                       & Pathological separation of the retinal neuroepithelium and the pigment epithelium (\cite{ghazi2002pathology}) \\
Myopia                           &    2251                       & The main feature is axial lengthening; high myopia may cause characteristic fundus degenerative changes (\cite{morgan2012myopia}) \\
Macular Scar                     &    1937                       & Fibrotic scar tissue formed after repair of inflammation, hemorrhage, or trauma in the macula (\cite{trincao2024gene}) \\
Glaucoma                         &    2280                       & An optic neuropathy with pathological intraocular pressure as the main risk factor, leading to optic atrophy and visual field loss (\cite{sahu2024review}) \\
Optic Disc Edema                 &    762                       & Abnormal blurred borders, bulging, thickening, pale or congested color, often with bleeding or exudation, indicating intracranial pressure or optic neuropathy (\cite{van2007optic}) \\
Diabetic Retinopathy             &    3444                       & Microvascular complications of diabetes, seen as retinal microaneurysms, hemorrhage, exudation, neovascularization, etc. (\cite{kour2024review}) \\
Central Serous Chorio retinopathy &    606                      & Localized serous detachment of the neuroepithelium in the macular area due to dysfunction of the retinal pigment epithelium (\cite{fung2023central}) \\
\bottomrule
\end{tabular}
\caption{Description of healthy and various fundus diseases and number of fundus images}
\label{tab:various fundus diseases}
\end{table}

\subsubsection{Data preprocessing}

We first examine the dataset by counting the number of fundus images per category and verifying the size of images. The size of all images belonging to this dataset is [2004, 1690] pixels. Because each of image in this dataset is rectangular, in order to optimize the input size and help the model convergence, the short side of all images is scaled to 256 pixels and the aspect ratio is kept unchanged; then the area with a pixel size of [224,224] is cropped from the center of the image, the input image is standardized, and then the image is converted to tensor format, and the pixel value of the image is scaled to the [0,1] interval. Then we calculate the mean and standard deviation of the images in this dataset and then normalize the images based on the mean and standard deviation. During the training process, random shuffling is used to enhance the generalization ability of the model. The test set and validation set do not use random shuffling to maintain the consistency of the evaluation, and multi-threaded loading is used to improve data reading efficiency.

\subsection{SwinECAT}

\begin{figure}[H]
    \centering
    \includegraphics[width=1.0\linewidth, keepaspectratio]{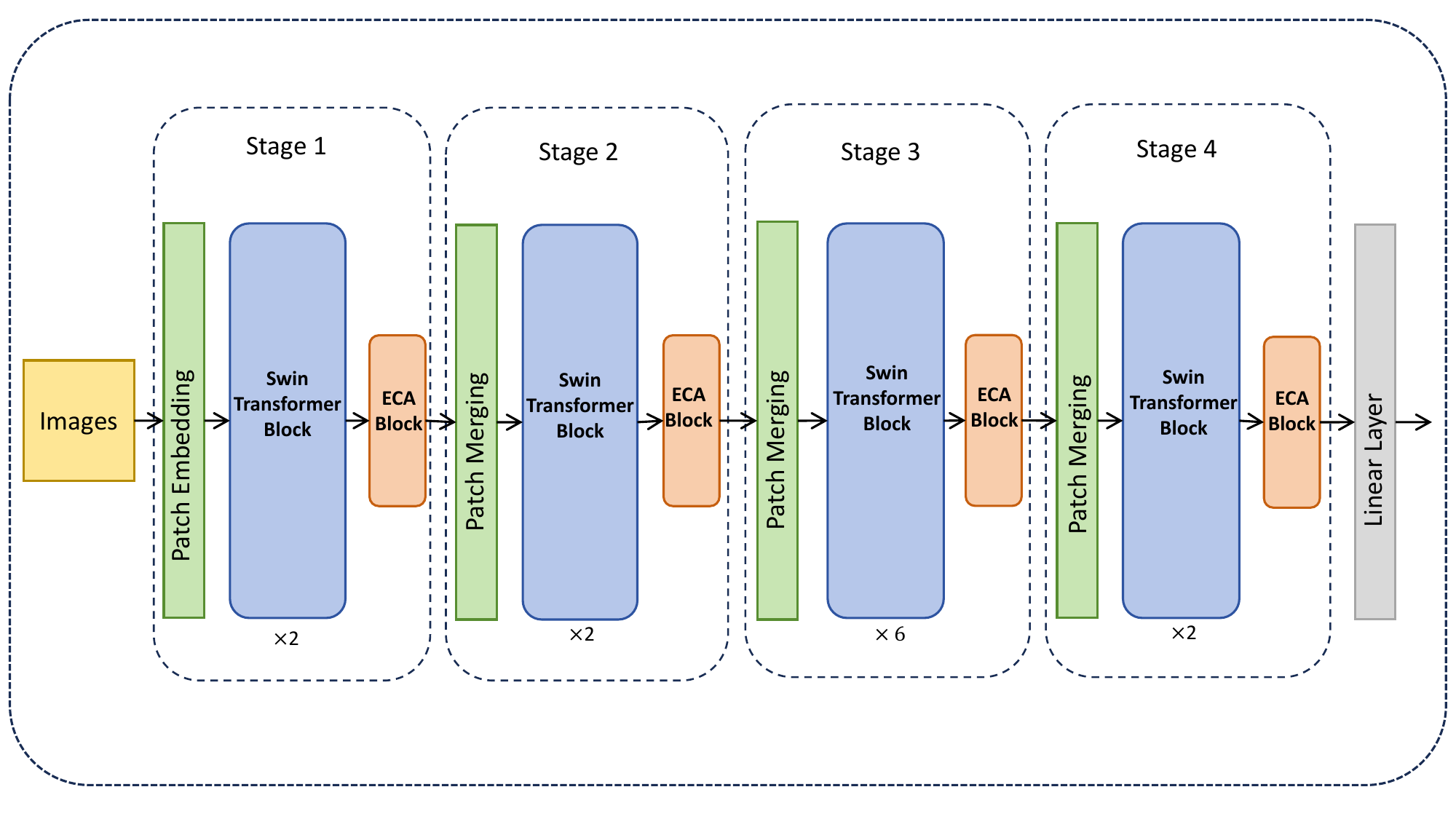}
    \caption{The overall structure of the SwinECAT model. It mainly consists of four stages, each of which mainly includes some Swin Transformer Blocks and an ECA Block.}
    \label{SwinECAT_1}
\end{figure}

\begin{figure}[H]
    \centering
    \includegraphics[width=1.0\linewidth, keepaspectratio]{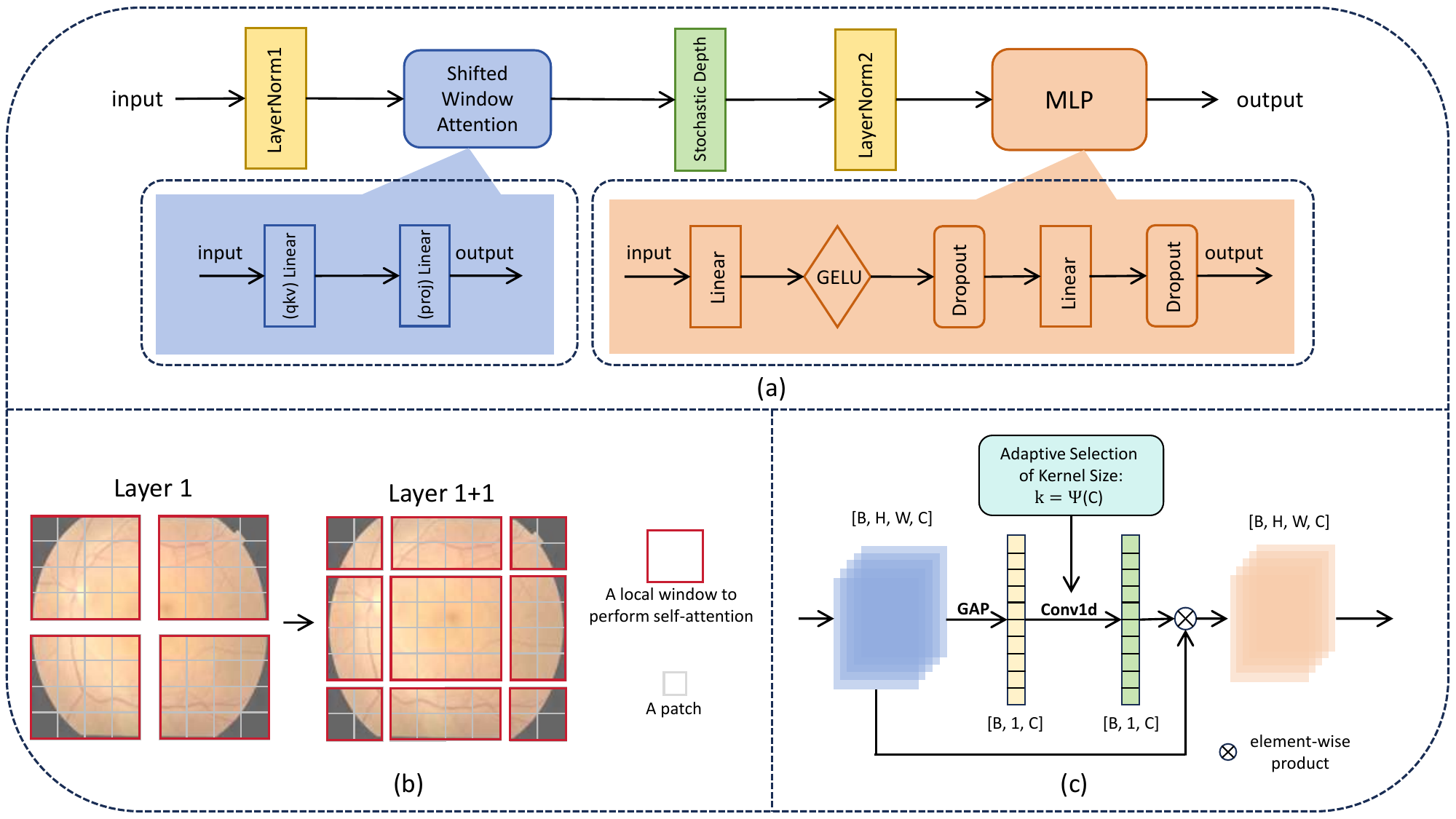}
    \caption{The key components of the SwinECAT model. Figure 2(a) shows the structure of the Swin Transformer Block, Figure 2(b) shows the shift window mechanism, and Figure 2(c) shows the local cross-channel interaction process of ECA's global average pooling and one-dimensional convolution.}
    \label{SwinECAT_2}
\end{figure}

As illustrated in Figures \ref{SwinECAT_1} and \ref{SwinECAT_2}, we present our proposed SwinECAT model from both the overall architecture and key components perspectives.

\subsubsection{Overall Structure of the SwinECAT model}

As shown in the figure ~\ref{SwinECAT_1} above, the SwinECAT model we proposed gradually builds a multi-scale feature map through a four-stage hierarchical design, that is, gradually reducing the spatial resolution and increasing the number of channels to extract multi-scale features. Similar to the backbone of Swin Transformer (~\cite{liu2021swin}) , it can effectively capture spatial information at different resolutions. The core of each stage of SwinECAT is several Shifted Window (Swin) Transformer Blocks and one Efficient Channel Attention (ECA) module (~\cite{wang2020eca}) . Stages 1, 2, and 4 all have two Swin Transformer Blocks and one ECA module, and stage 3 has six Swin Transformer Blocks and one ECA module. This is because the resolution of stage 3 is low, the computational cost is low, and it is suitable for deepening the network structure, which helps to improve the feature representation ability. In Stage 4, the receptive field is already sufficiently large, so adding more Swin Transformer Blocks yields diminishing returns while increasing the risk of overfitting. Therefore, stage 4 only uses two Swin Transformer Blocks and one ECA module. This hierarchical feature expression enables the model to take into account both local details and global context, improving visual understanding capabilities. On this basis, ECA module is added after the four stages to further enhance the model's feature expression capabilities.

The SwinECAT model we proposed maintains the advantages of the hierarchical feature map of the backbone of Swin Transformer. Unlike the original Vision Transformer (ViT) which treats images as flat sequences of a single scale, the Swin transformer constructs a pyramid-like multi-scale feature map similar to Feature Pyramid Network (~\cite{lin2017feature,alzubaidi2021review,dosovitskiy2020image,liu2021swin}) . This hierarchical structure enables it to efficiently capture different scales of information in the image, from local details to global semantics, which is crucial for the correct classification of fundus images.

\subsubsection{Structure of the Swin Transformer Block}

The Swin Transformer Block is the core building block of the SwinECAT model, similar to the Swin Transformer baseline model. As shown in Figure \ref{SwinECAT_2}(a), its Shifted Window Attention block contains the window self-attention and shifted window self-attention mechanisms, which perform self-attention calculations in local windows and implement cross-region information interaction through shift operations between windows. In addition to the self-attention layer, it also contains an MLP layer, which is mainly composed of two fully connected layers and a nonlinear activation function GELU, which is used to further process the output of self-attention and enhance the expressiveness of the model.As shown in Figure \ref{SwinECAT_2}(b), the model first divides the image into non-overlapping local windows, calculates self-attention within each window, and regularly shifts the window divisions in the next layer, so that adjacent windows contain information from different windows in the previous layer. Cross-window connection cleverly introduces information interaction between adjacent windows and solves the problem of information isolation between windows caused by calculating attention within a fixed window.

The SwinECAT model adopts the window self-attention mechanism and shifted window self-attention mechanism of Swin Transformer backbone, which takes into account both local and global modeling capabilities while maintaining computational efficiency. It is suitable for processing the fine-grained and widely distributed features of lesions in fundus images (~\cite{liu2021swin}) .

\subsubsection{Structure of the efficient channel attention}

The core goal of ECA is to significantly improve the feature expression capability of neural networks without increasing the amount of computation. As shown in Figure \ref{SwinECAT_2}(c), it captures the local dependencies between channels through simple and efficient one-dimensional convolution operations, and realizes adaptive adjustment of the weights of each channel, so that the model can pay more attention to key information channels and suppress irrelevant or redundant features. Compared with previous channel attention mechanisms such as Squeeze-and-excitation (SE) (~\cite{hu2018squeeze}) , ECA has the advantages of fewer parameters and low computational overhead, and is suitable for integration into multi-scale feature maps without significantly increasing the complexity of the model (~\cite{wang2020eca}) . It is an efficient and lightweight channel attention module. 

If the input image is $\mathbf{X} \in \mathbb{R}^{H \times W \times C}$ , the main calculation process of the ECA module can be expressed as the following three steps:

The spatial position of each channel c is averaged to obtain the channel description vector:

\begin{equation}
z_c = \frac{1}{H \times W} \sum_{i=1}^{H} \sum_{j=1}^{W} X_{i, j, c}, \quad \forall c \in [1, C]
\end{equation}

If $Conv1D$ is a convolution operation sliding on the channel dimension, where the kernel size $k$ is adaptively determined based on the number of channels $C$, and $\sigma$ is the Sigmoid activation function used to map the output to $[0,1]$, then one-dimensional convolution is used to model the channel dimension and generate channel attention weights:

\begin{equation}
\mathbf{w_c} = \sigma \left( \text{Conv1D} (\mathbf{z_c,k}) \right) 
\end{equation}

Finally, multiply the attention weight w back to the corresponding channel:

\begin{equation}
\hat{X}_{i,j,c} = w_c \times X_{i,j,c}
\end{equation}

After applying the ECA module to the multi-scale features of Swin transformer backbone , it strengthens the complementary fusion of the self-attention mechanism in the spatial dimension and the channel dimension. So SwinECAT may improve the ability of the Swin Transformer baseline model to distinguish small category differences, and improves the performance of Swin Transformer baseline model in medical image classification, especially for fundus images with rich details. At the same time, the adaptive weighting mechanism of ECA helps to improve the model's generalization ability, so that the model can still maintain stable performance under different data distributions.

We improve the backbone of Swin Transformer (~\cite{liu2021swin}) by introducing an ECA module (~\cite{wang2020eca}) after each stage of its four stages for downsampling and channel expansion, making full use of the spatial and channel attention mechanisms to help improve its performance on vision tasks.

\subsection{Model evaluation metrics}

This study evaluates the models on the test set using accuracy, macro-averaged and weighted-averaged precision, recall, and F1 scores, aiming to comprehensively assess model performance from multiple perspectives and mitigate the limitations of relying on a single metric. Accuracy measures the overall prediction correctness and provides a basic accuracy reference. The macro-average equally weights the indicators of each category to avoid the minority class being masked by the majority class and ensure category fairness. The weighted-average is calculated based on the proportion of category samples, which fits the actual data distribution and reflects the performance of the model in real scenarios. The goal of this is to effectively make up for the shortcomings of a single indicator in terms of category imbalance and scene adaptation, and to fully reveal the advantages and disadvantages of the models.

\section{Results and discussion}

The experimental results primarily include comparisons between our proposed model SwinECAT and widely used visual baseline models, as well as baseline models from previous studies on fundus disease image classification. There are also ablation experiments between SwinECAT and the Swin Transformer baseline model. The ablation experiment verifies the effectiveness of our proposed swinECAT. Compared with the methods proposed in previous studies, SwinECAT has fewer model parameters and better results.

\subsection{Experimental setup}

We use the EDID augmented dataset for model training and evaluation. The dataset includes 16,140 images in 9 categories. We divide 80\% of the dataset into training sets, 10\% of the dataset into validation sets, and 10\% of the dataset into test sets.

In the experimental training process, each model uses the cross entropy loss function as the objective function of the classification task, and the model parameters are optimized by the Adam optimizer. The batch size of the training set, validation set, and test set is 32, and the learning rate is set to 1e-5. In order to prevent the model from overfitting and reduce unnecessary training costs, we use the early stopping mechanism. Each model uses the early stopping mechanism to select the best training results. If the validation set loss value fails to decrease in three consecutive rounds of training, the model training is terminated and the model parameters with the smallest validation loss are retained as the final model.

\subsection{Experimental results}

In the experimental part, the SwinECAT model we proposed is compared with several widely used baseline models for general vision tasks, including five Transformer-based models and one CNN-based model ResNet50 (~\cite{he2016deep}) . These five Transformer-based visual models are: ViT (~\cite{dosovitskiy2020image})  Swin Transformer (~\cite{liu2021swin}) , BEiT (~\cite{bao2021beit}) , ConViT (~\cite{d2021convit}) , and MaxViT (~\cite{tu2022maxvit}) . In addition, we design comparative experiments between SwinECAT and recently proposed methods for fundus disease image classification, including the feature fusion model of MaxViT and ResNet (~\cite{liu2024automated}) , and the sequential connection model of CNN and Transformer (~\cite{lian2024lesion}) . The feature fusion model of MaxViT and ResNet is to fuse the final output features of the MaxViT and ResNet18, and then classify them according to the fused feature results (~\cite{liu2024automated}) . The sequential connection model of CNN and Transformer is to first use the CNN-based model Inception-ResNet-v2 to extract features, then process the output dimensions, and input them into the ViT model for further processing (~\cite{lian2024lesion}) . The comparison between the SwinECAT model and the backbone of Swin Transformer is the ablation experiment designed in this paper to prove the effectiveness of the ECA module designed in this paper.

\subsubsection{Experimental results of the SwinECAT model and other baseline models}

As shown in the following Figure \ref{fig:Figure Result}, the training process of each model mainly includes the loss function and accuracy change of the training set and the validation set. During the model training process, the model training hyperparameters such as batch size, learning rate, and the optimizer are kept consistent. All models use the same early stopping mechanism to select the better experimental results.

\begin{figure}[H]
  \centering

  \begin{subfigure}[b]{0.3\textwidth}
    \centering
    \includegraphics[width=\linewidth]{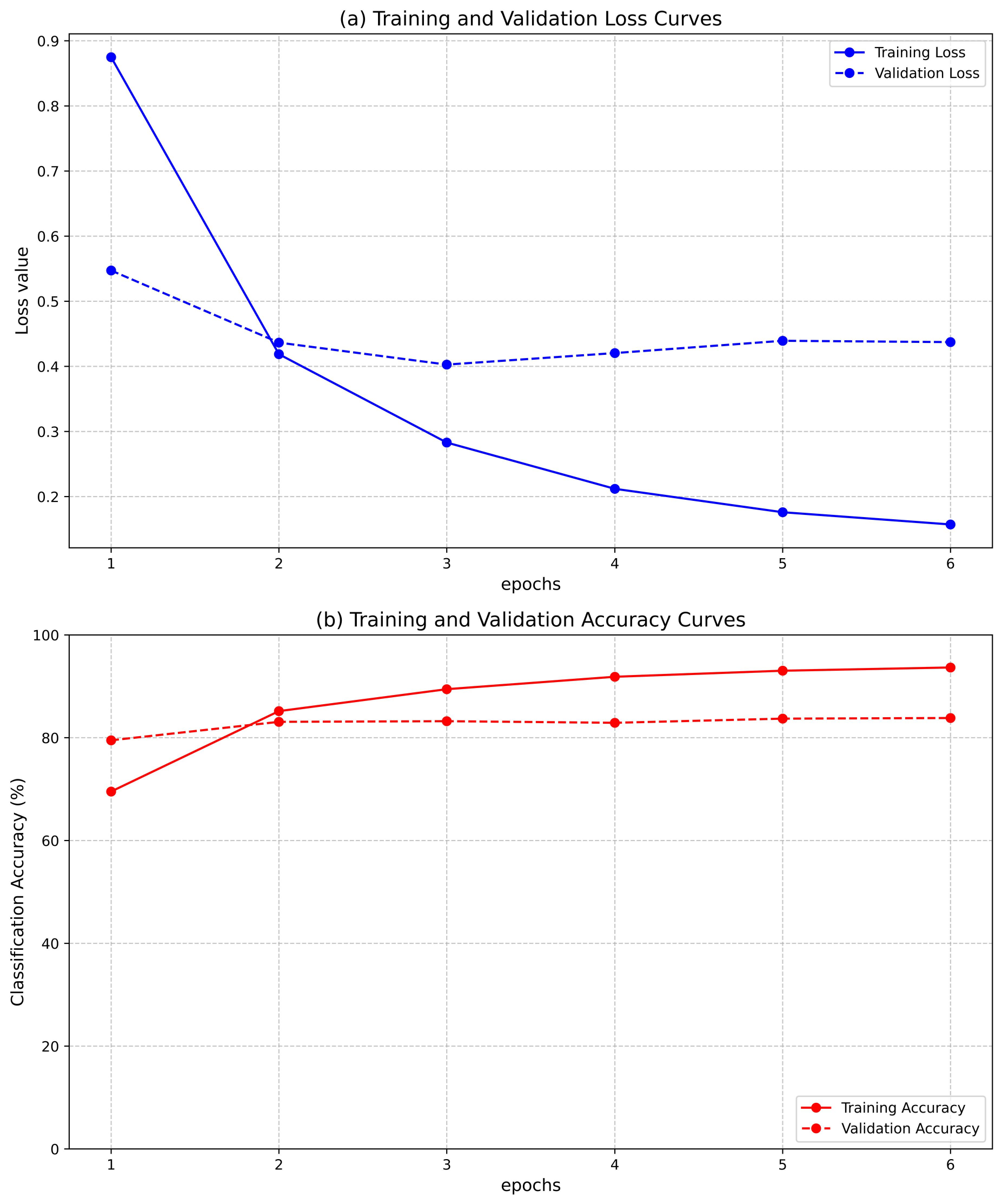}
    \caption{(a) ViT}
  \end{subfigure}
  \begin{subfigure}[b]{0.3\textwidth}
    \centering
    \includegraphics[width=\linewidth]{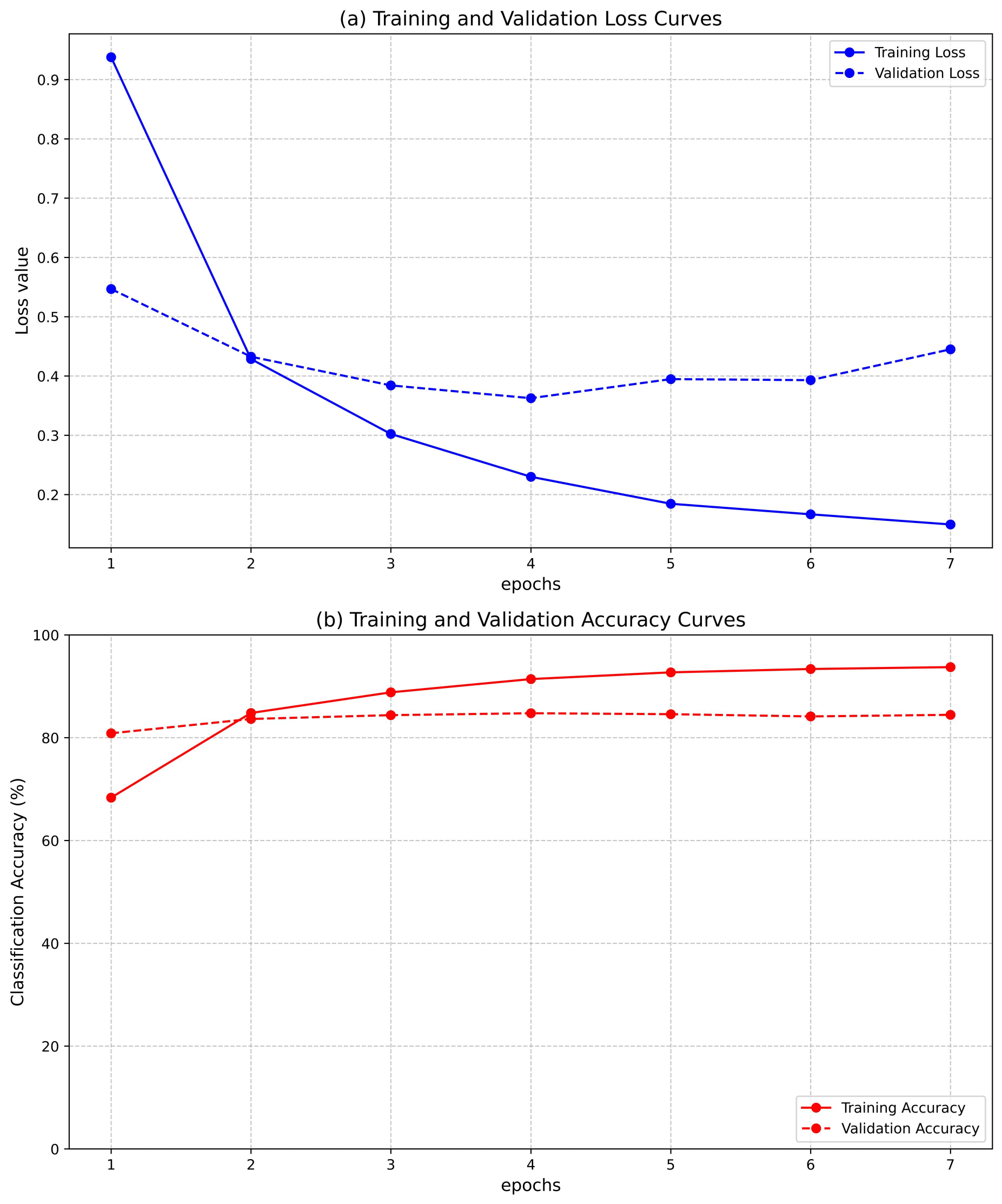}
    \caption{(b) ConViT}
  \end{subfigure}
  \begin{subfigure}[b]{0.3\textwidth}
    \centering
    \includegraphics[width=\linewidth]{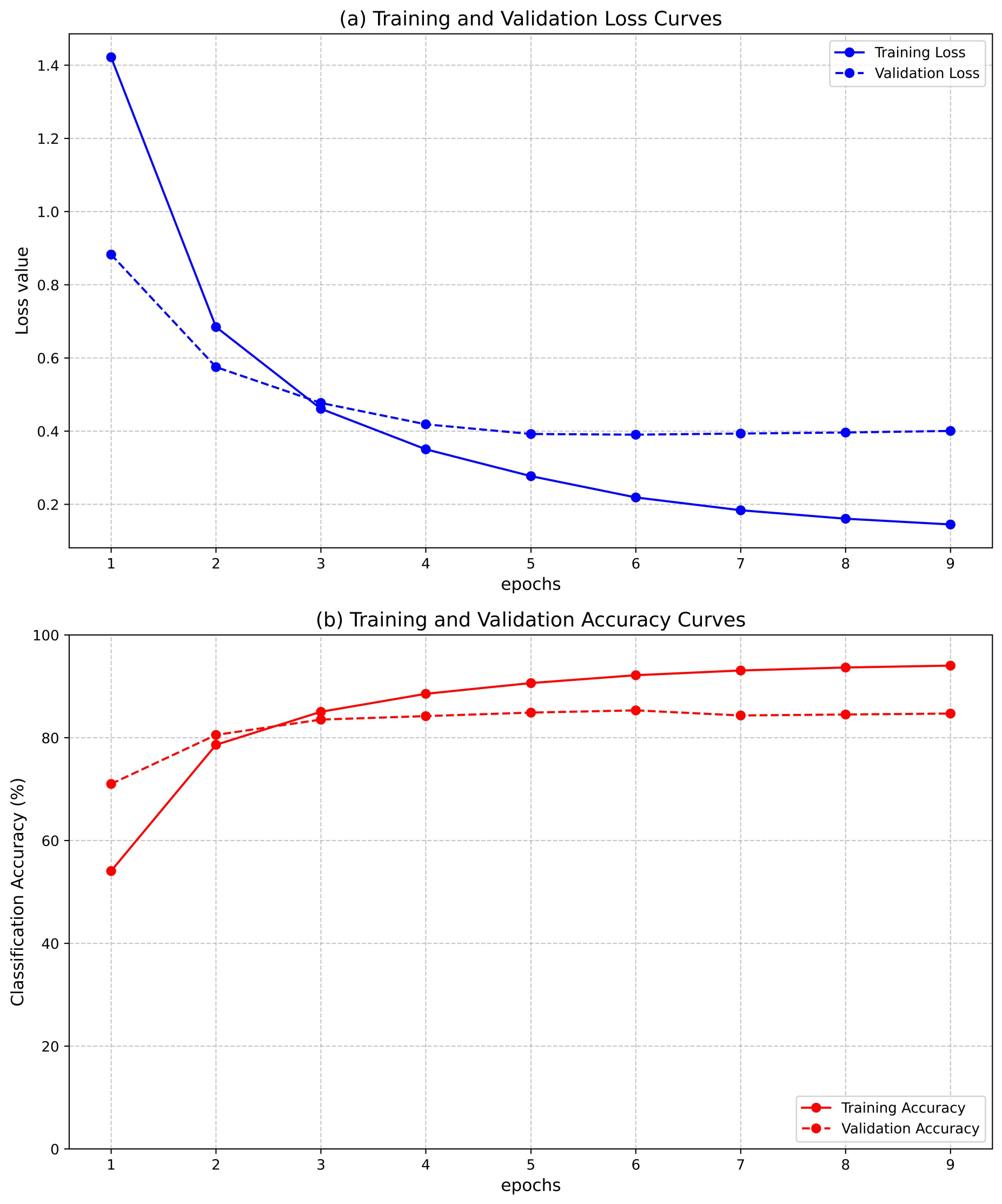}
    \caption{(c) MaxViT}
  \end{subfigure}

   \vspace{0.1cm}

  \begin{subfigure}[b]{0.3\textwidth}
    \centering
    \includegraphics[width=\linewidth]{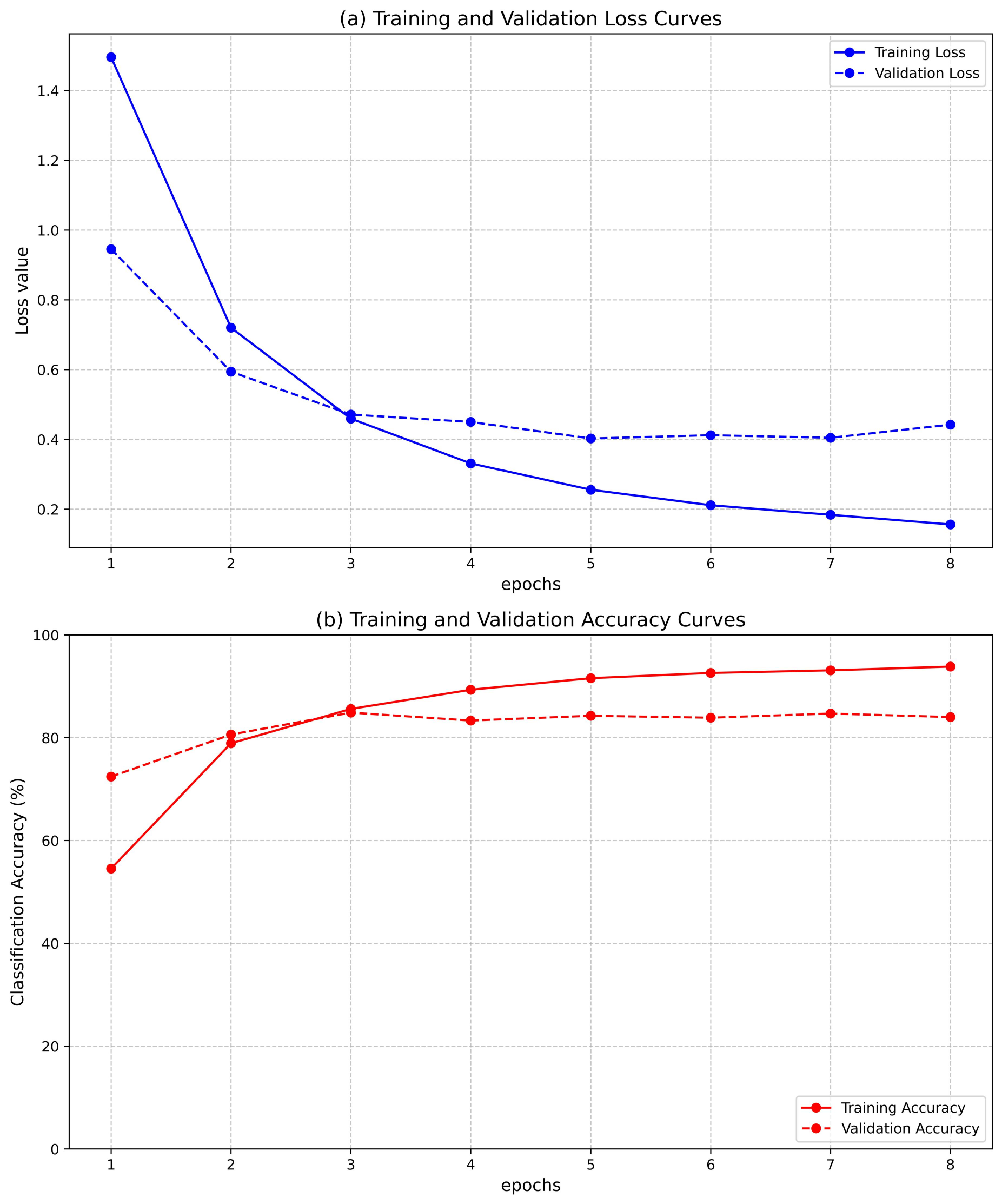}
    \caption{(d) BEiT}
  \end{subfigure}
  \begin{subfigure}[b]{0.3\textwidth}
    \centering
    \includegraphics[width=\linewidth]{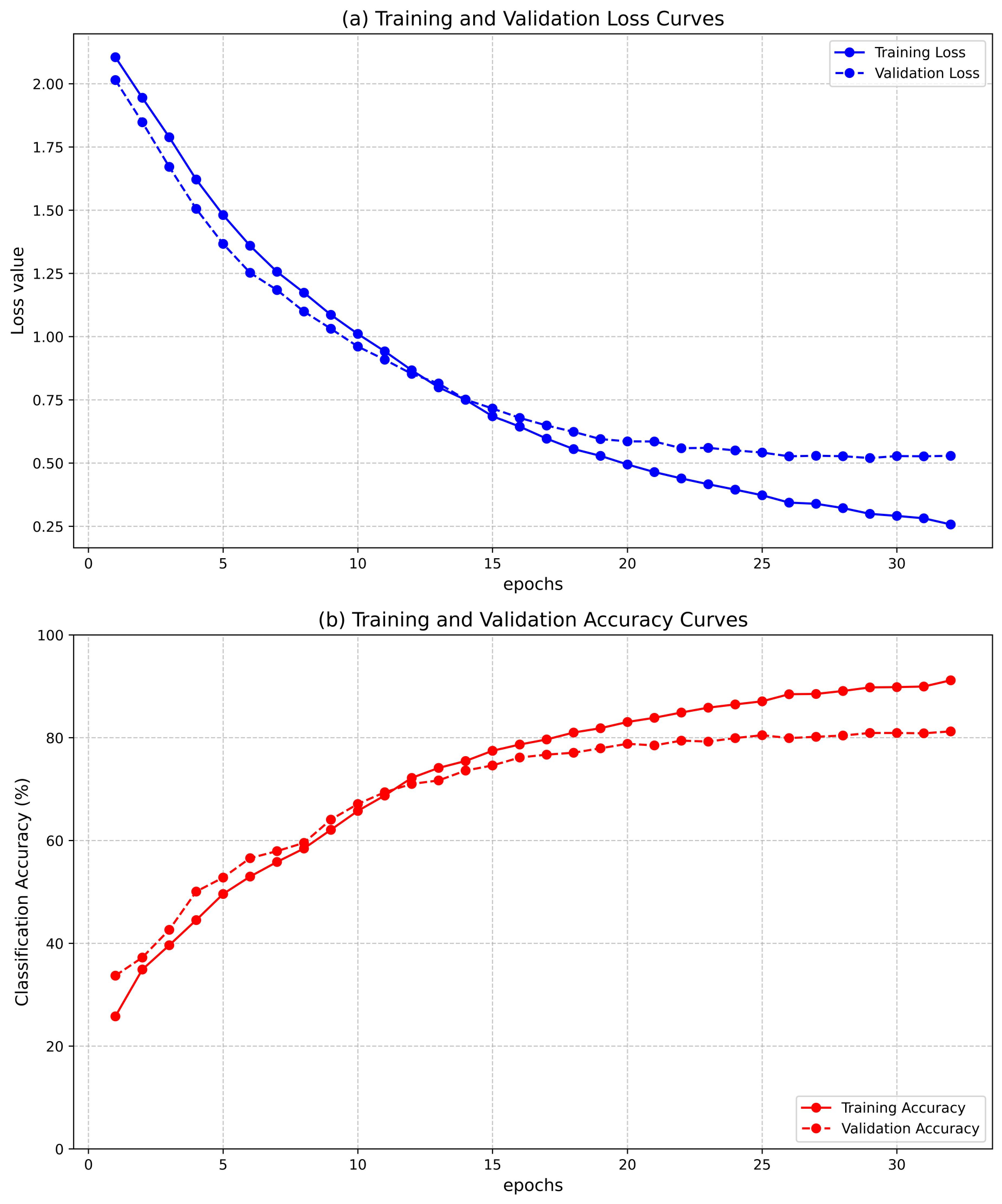}
    \caption{(e) ResNet50}
  \end{subfigure}
  \begin{subfigure}[b]{0.3\textwidth}
    \centering
    \includegraphics[width=\linewidth]{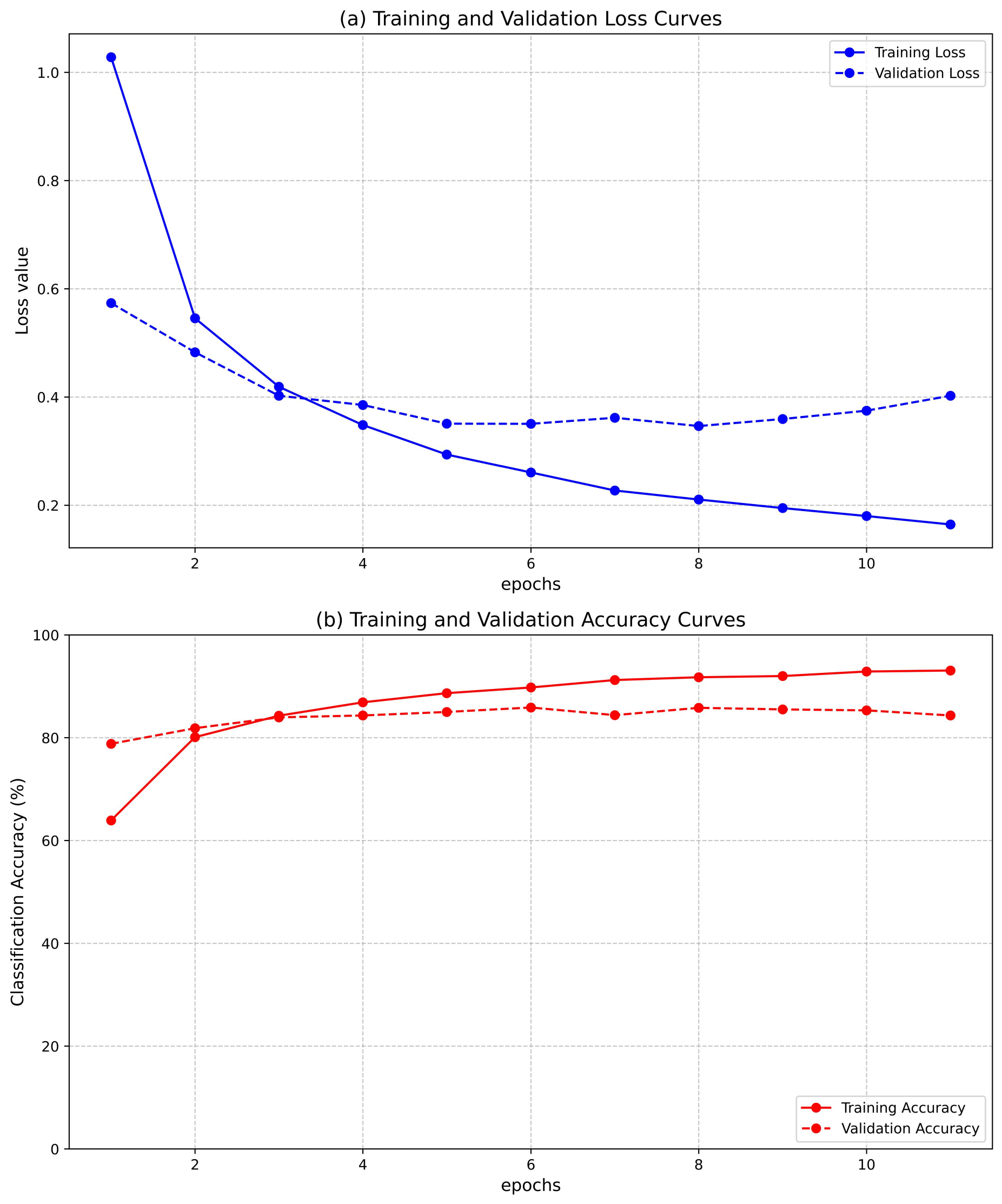}
    \caption{(f) Swin Transform}
  \end{subfigure}
  
   \vspace{0.1cm}

  \begin{subfigure}[b]{0.3\textwidth}
    \centering
    \includegraphics[width=\linewidth]{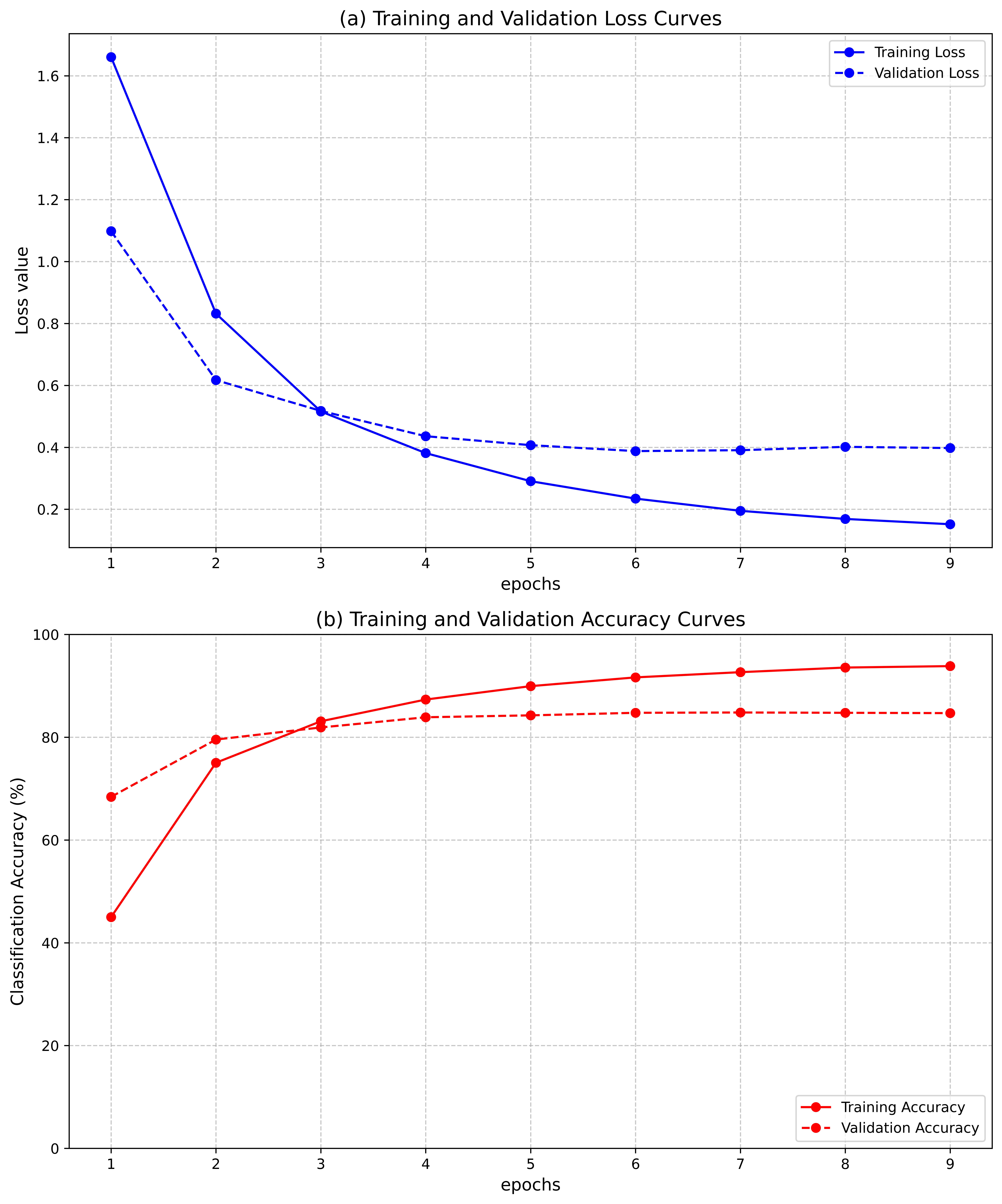}
    \caption{(g) MaxViT+ResNet }
  \end{subfigure}
  \begin{subfigure}[b]{0.3\textwidth}
    \centering
    \includegraphics[width=\linewidth]{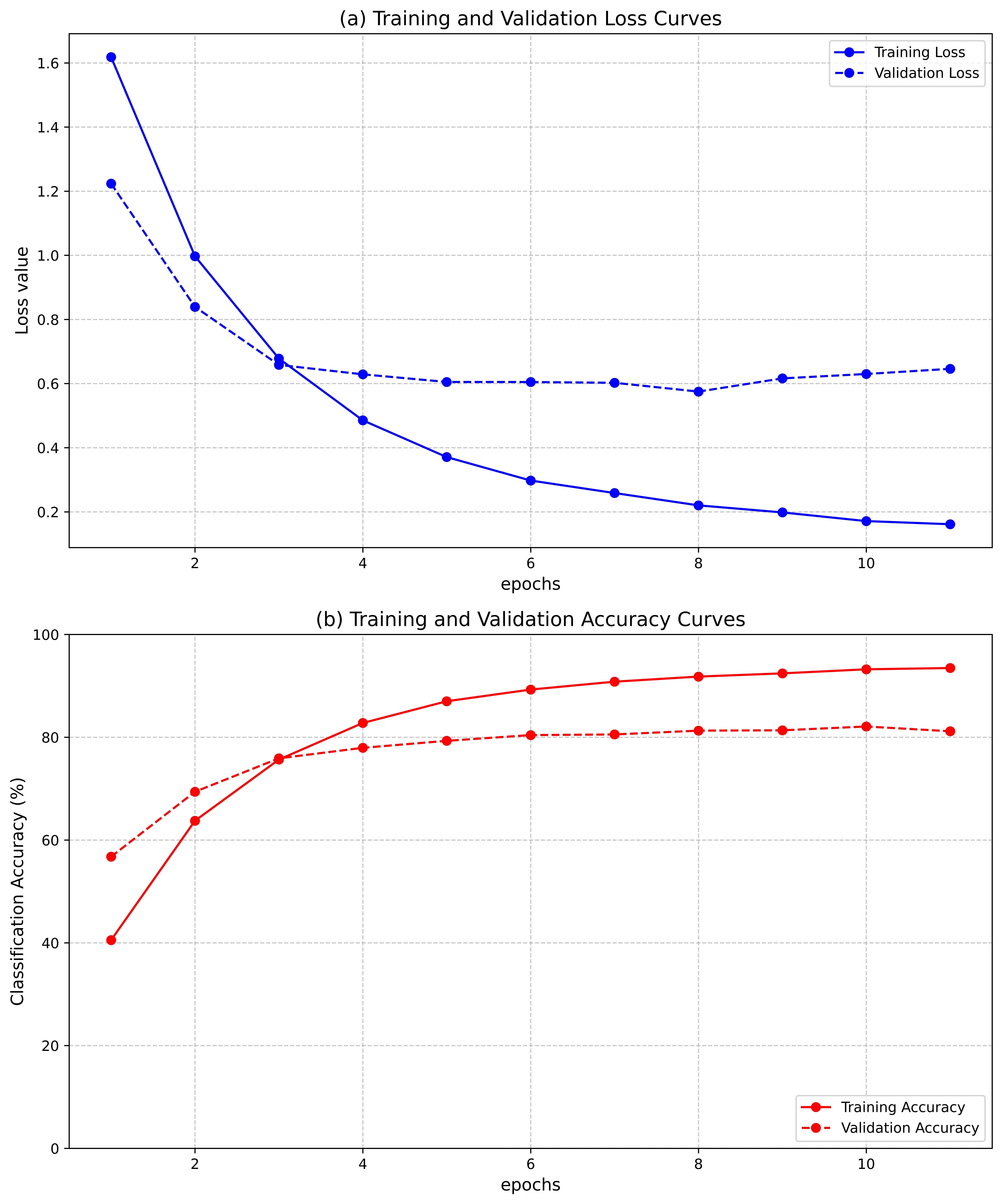}
    \caption{(h) CNN-Transformer}
  \end{subfigure}
  \begin{subfigure}[b]{0.3\textwidth}
    \centering
    \includegraphics[width=\linewidth]{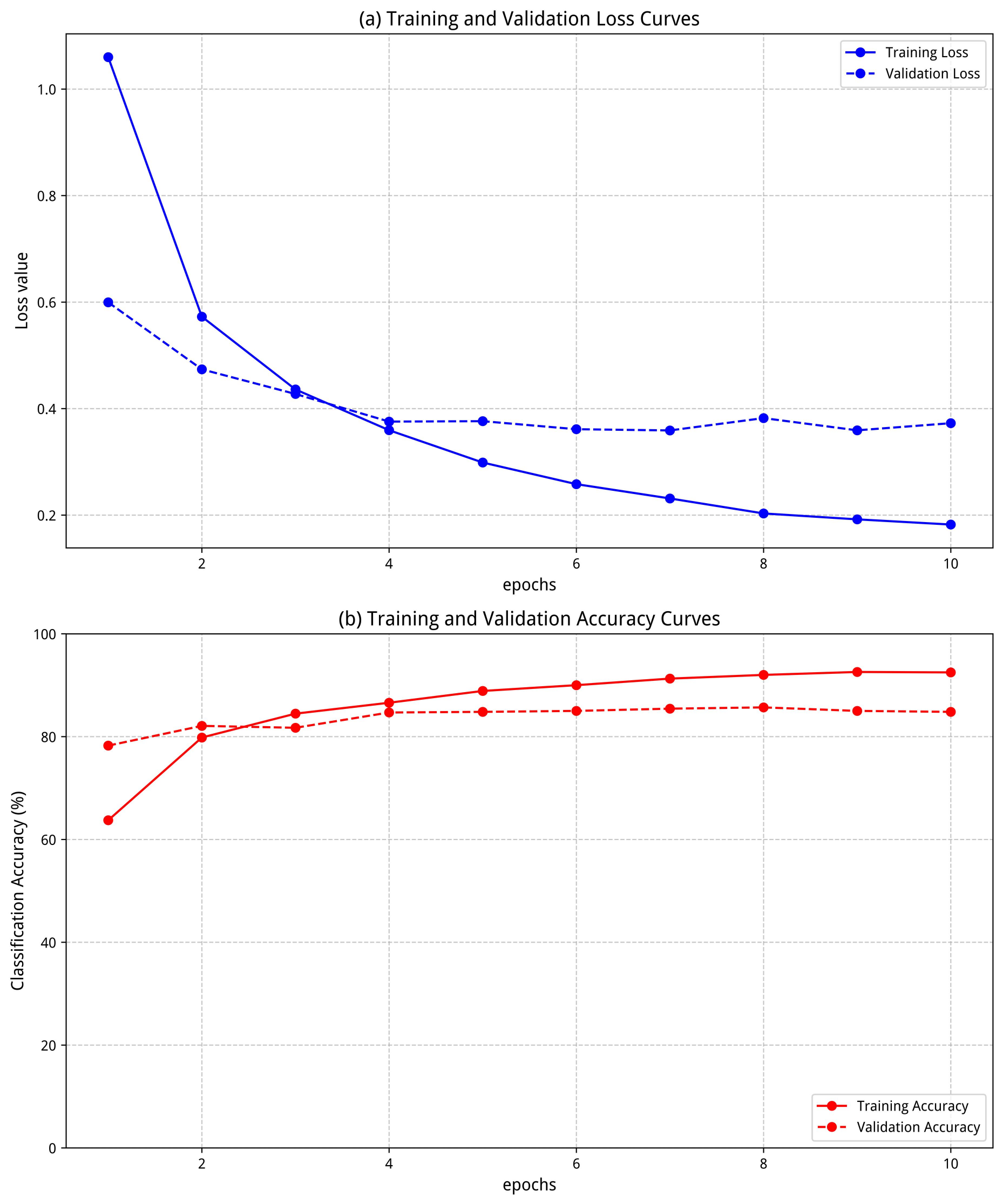}
    \caption{(i) SwinECAT}
  \end{subfigure}
  
  \caption{Loss function and accuracy of SwinECAT and other comparison models during training process. Figure 3(a-f) shows the training process of each widely used baseline model for general vision tasks, Figure 3(g) shows the training process of the feature fusion model of MaxViT and ResNet, Figure 3(h) shows the training process of the sequential connection model of CNN and Transformer, and Figure 3(i) shows the training process of the SwinECAT model.}
  \label{fig:Figure Result}
  
\end{figure}

Through the above experiments, as shown in Figures \ref{fig:Figure Result}(a-f), the loss rate of Swin Transformer on the validation set is smaller than that of other baseline models, indicating that the backbone of Swin Transformer fits the objective function better on this task. As shown in Figure \ref{fig:Figure Result}(i), the SwinECAT model we proposed alleviates the rising trend of the loss rate of Swin Transformer in the later rounds to a certain extent. As shown in Figure \ref{fig:Figure Result}(g-i), the loss rate of the SwinECAT model we proposed on the validation set is always smaller than the combined model method of Figure \ref{fig:Figure Result}(g) and Figure \ref{fig:Figure Result}(i).

When the loss of a model is smaller than that of other models, it usually means that it fits the data better under the current loss function, but this does not necessarily mean that its overall performance is better. The model can be considered to perform better when the main evaluation indicators such as accuracy and F1 score are also better.As shown in Table \ref{Table:Results}, the accuracy, macro-average and weighted average precision, recall, and F1 score of SwinECAT and each baseline model on the test set, as well as the number of parameters of each model.

\begin{table}[H]
\centering
\resizebox{\textwidth}{!}{  
\begin{tabular}{ccccccccc}
\hline
Models & Accuracy(\%) & Macro Precision(\%) & Weighted Precision(\%) & Macro Recall(\%) & Weighted Recall(\%) & Macro F1 & Weighted F1 & Parameters(M) \\
\hline
ViT (~\cite{dosovitskiy2020image})               &  85.01        &  88.64        &   85.42       &    86.67      &   85.01       &      0.8752    &   0.8510       & \textbf{21.67} \\
ConViT (~\cite{d2021convit})           &   86.06      &    88.56      &    86.15      &    87.69      &    86.06      &          0.8799 &    0.8597      & 27.35 \\
MaxViT (~\cite{tu2022maxvit})           &  86.49        &    89.38      &    86.76      &    88.01      &   86.49       &      0.8859    &    0.8653      & 30.41 \\
BEiT (~\cite{bao2021beit})             &    85.32      &     88.27     &    85.48      &     85.87     &   85.32       &       0.8691   &     0.8530     & 85.77 \\
ResNet50 (~\cite{he2016deep})         &   83.27      &     85.61     &   83.33       &    83.37      &     83.27     &       0.8426   &    0.8319      & 23.53 \\
Swin Transformer (~\cite{liu2021swin}) &   86.56       &    88.70     &    86.58      &    88.38     &    86.56      &       0.8849   &     0.8654     & 27.53 \\
MaxViT+ResNet18 (~\cite{liu2024automated})  &   87.42     &    89.74              &    87.82               &     89.22          &     87.42           &     0.8936      &   0.8774         &      42.11      \\
CNN-Transformer (~\cite{lian2024lesion}) &  79.93      &    83.58              &   80.69                &     80.32          &     79.93           &    0.8160       &    0.8000        &    76.07        \\
\textbf{SwinECAT}             &    \textbf{88.29}      &      \textbf{90.82}            &   \textbf{88.55}                &     \textbf{89.37}          &   \textbf{88.29}             &     \textbf{0.9000}      &   \textbf{0.8830}         &    28.30        \\
\hline
\end{tabular}
}
\caption{Comparison of the performance on the test set of our proposed SwinECAT model with various baseline models}
\label{Table:Results}
\end{table}

The comparative results between SwinECAT and six popular general baseline models are presented in the first 6 rows of Table \ref{Table:Results}. Rows 7 and 8 of Table \ref{Table:Results} show the comparative results between SwinECAT and two recently proposed methods for fundus disease classification: the feature fusion model combining MaxViT and ResNet, and the sequential connection model of CNN and Transformer.

The SwinECAT we proposed outperforms other baseline models in terms of accuracy, precision, recall, F1 scores model average indicators. The model accuracy reached 88.29\%, the macro-average and weighted average precision reached 90.82\% and 88.55\% respectively, the macro-average and weighted average recall reached 89.37\% and 88.29\% respectively, and the macro-average and weighted average F1 scores were 0.9000 and 0.8830 respectively. On the unbalanced dataset EDID, the SwinECAT model achieved high performance in both macro-average and weighted average indicators, indicating that it can not only accurately identify common fundus disease types with a high proportion in the dataset, but also has good recognition capabilities for fundus disease types with fewer samples. 

As shown in the first 6 rows of Table\ref{Table:Results}, our proposed SwinECAT model outperforms six widely used general vision baseline models in the task of fundus disease image classification. Among these 6 baseline models, Swin Transformer achieved better results than other models in terms of comprehensive accuracy and model parameters. The accuracy of the Swin Transformer base model reached 86.56, the highest among the 6 base models. Compared with MaxViT, which ranks second in accuracy, Swin Transformer has fewer parameters, a simpler model structure, and is easier to improve. Therefore, we improved the backbone model of Swin Transformer to make it more suitable for fundus images, and proposed the SwinECAT model.

As shown in rows 7-8 of Table \ref{Table:Results}, SwinECAT performs better and has fewer parameters than two baseline models proposed in recent studies on fundus disease classification tasks. The feature fusion model of MaxViT and ResNet (~\cite{liu2024automated}) achieves an accuracy of 87.42\% on the test set, which is only lower than our proposed SwinECAT model among the nine models in Table \ref{Table:Results}, but the number of parameters is 1.5 times that of our proposed SwinECAT model. The sequential connection model of CNN and Transformer (~\cite{lian2024lesion}) achieves an accuracy of only 79.93\% on the test set, and has a large number of parameters, reaching 76.07M.

\subsubsection{Ablation experiment}

In order to demonstrate the effectiveness of the ECA module in our designed SwinECAT model, we designed corresponding ablation experiments. That is, the SwinECAT model is compared with the Swin Transformer baseline model, and the training hyperparameters are kept consistent. Both models use the same early stopping mechanism.

As shown in rows 6 and 9 of Table \ref{Table:Results}, the SwinECAT model we proposed has improved all indicators compared to the Swin Transformer baseline model without increasing the number of model parameters too much. Swin Transformer effectively models spatial structural information through local window attention, while the ECA module further enhances the model's ability to selectively focus on key feature channels, thereby improving the discriminability of feature representation. This makes the model more sensitive in extracting structural features of fundus images, and ultimately achieves higher performance in this task.

\subsection{Discussion}

Based on the experimental results shown in Table \ref{Table:Results} and Figure \ref{fig:Figure Result}, we discuss the possible reasons for this result. The superior performance of the Swin Transformer baseline model compared to the other five general vision models may be attributed to its three core advantages. That is, the multi-scale feature representation of the pyramid structure of the backbone of Swin Transformer, the window self-attention mechanism, and the shifted window self-attention mechanism. These mechanisms and characteristics may be more in line with the characteristics of fundus images with small details and complex structures, and can extract local details and global semantics of fundus images.

Compared with the global self-attention mechanism of ViT (~\cite{dosovitskiy2020image}) , BEiT (~\cite{bao2021beit}) and ConViT (~\cite{d2021convit}) , the local attention mechanism of the SwinECAT model designed by us based on the Swin Transformer backbone (~\cite{liu2021swin}) can focus on key local areas more efficiently, and the ECA mechanism helps to select more critical feature channels. This paper speculates that models such as ViT, BEiT, ConViT, etc., which mainly use the global attention mechanism, are difficult to fully model local features when facing fundus images with complex distribution of lesion areas, and therefore perform worse than SwinECAT. Although MaxViT (~\cite{tu2022maxvit}) integrates local and global attention mechanisms, its model structure is more complex and has a little more parameters than SwinECAT, and its performance on this task is relatively inferior to SwinECAT. Overall, SwinECAT has achieved an improvement over the general visual baseline model Swin Transformer in the task of fundus disease image classification while maintaining parameter efficiency, and achieved better performance on this task.

Compared with recently proposed fundus disease classification models, such as the feature fusion model of MaxViT and ResNet18 (~\cite{liu2024automated}) and the sequential connection model of CNN and Transformer (~\cite{lian2024lesion}) , the use of ECA (~\cite{wang2020eca}) in the backbone of the Swin Transformer offers a lighter and more efficient improvement. Compared with the feature fusion model of MaxViT and ResNet18, which performs better among the two baseline methods, our proposed SwinECAT model reduces the number of parameters by nearly one-third while achieving 0.8\% improvement in accuracy. This may be due to the fact that both the feature fusion model and the sequential connection model have complex architectures and a large number of parameters, which makes them more prone to overfitting. In contrast, SwinECAT maintains the simplicity of the architecture, mainly through the channel attention mechanism to enhance the backbone of Swin Transformer's ability to select key channels, reducing the risk of overfitting caused by complex fusion.

\section{Conclusion}

In this work, we propose a Transformer-based architecture named SwinECAT, which effectively combines the Shifted Window Attention mechanism (\cite{liu2021swin}) with the Efficient Channel Attention mechanism (\cite{wang2020eca}). This hybrid design aims to enhance the model’s capability for fundus disease image classification. The Shifted Window Attention mechanism in SwinECAT restricts self-attention computation to local windows and enables cross-region information exchange through window shifting between layers. This design facilitates the capture of both local and global contextual information while maintaining computational efficiency. In addition, the integration of the ECA module allows the network to adaptively adjust channels, enabling it to focus more effectively on the most informative features. This synergistic fusion of spatial and channel attention mechanisms effectively improves the performance of fundus disease image classification compared to the Swin Transformer baseline model.

To validate the effectiveness of the proposed SwinECAT model, we conducted comprehensive experiments on the EDID dataset (\cite{sharmin2024dataset}). The SwinECAT model was evaluated against six widely used baseline models for general vision tasks, including five Transformer-based models (~\cite{dosovitskiy2020image,liu2021swin,d2021convit,bao2021beit,tu2022maxvit}) and the ResNet50 (~\cite{he2016deep}) model. In addition to these baselines for general vision tasks, we further compared SwinECAT with recently proposed methods specifically designed for fundus disease image classification, such as the feature fusion model of MaxViT and ResNet (\cite{liu2024automated}) and the sequential connection model of CNN and Transformer (~\cite{lian2024lesion}). Experimental results demonstrate that SwinECAT consistently outperforms the baseline methods, including the Swin Transformer baseline model and other approaches, across multiple evaluation metrics such as accuracy, precision, and F1-score. These findings indicate that incorporating a lightweight channel attention mechanism into the Swin Transformer backbone effectively enhances classification performance for fundus image-based disease diagnosis, while introducing only a minimal increase in model complexity. This highlights the high practical value of SwinECAT in medical image analysis applications.

Our study achieved good prediction results on a single image modality. In future studies, we may consider fusing multiple modalities for classification, such as the patient's description or a full-view video of the fundus, which may further improve the classification accuracy of various fundus diseases.

\newpage
\printbibliography

\end{document}